\documentclass{mva_style}

\usepackage{epsfig}

\usepackage{calc}
\usepackage{amssymb}
\usepackage{amstext}
\usepackage{amsmath}
\usepackage{amsthm}
\usepackage{multicol}
\usepackage{pslatex}
\usepackage{apalike}

\usepackage[table]{xcolor}
\usepackage{amssymb}
\usepackage{pifont}
\usepackage{array, makecell} %
\usepackage{svg}
\usepackage{times}
\usepackage{epsfig}
\usepackage{graphicx}
\usepackage{rotating}
\usepackage{amsmath}
\usepackage{amssymb}
\usepackage{multirow}
\usepackage{diagbox}
\usepackage{subfig}
\usepackage{hyperref}

\newcommand{\cmark}{\ding{51}}
\newcommand{\xmark}{\ding{55}}

\finalcopy

\begin{document}


\title{Detecting Object States vs Detecting Objects:\\A New Dataset and a Quantitative Experimental Study}


\author{Filipos Gouidis$^{1,2}$, Theodore Patkos$^1$,  Antonis Argyros$^{1,2}$,Dimitris Plexousakis$^{1,2}$ \\
$^1$Institute of Computer Science, FORTH, Greece\\                 
$^2$Computer Science Department, University of Crete, Greece\\
{\tt\small \{gouidis,patkos,argyros,dp\}@ics.forth.gr}
} 


 \maketitle \normalsize 

\section*{\centering Abstract}
\textit{
The detection of object states in images (State Detection - SD) is a problem of both theoretical and practical importance and it is tightly interwoven with other important computer vision problems, such as action recognition and affordance detection. It is also highly relevant to any entity that needs to reason and act in dynamic domains, such as robotic systems and intelligent agents. Despite its importance, up to now, the research on this problem has been limited. In this paper, we attempt a systematic study of the SD problem. First, we introduce the \textit{Object State Detection Dataset (OSDD)}, a new publicly available dataset consisting of more than 19,000 annotations for 18 object categories and 9 state classes. Second,  using a standard deep learning framework used for Object Detection (OD), we conduct a number of appropriately designed experiments, towards an  in-depth study of the  behavior of the SD problem. This study enables the setup of a baseline on the performance of SD, as well as its relative performance in comparison to OD, in a variety of scenarios. Overall, the experimental outcomes confirm that SD is harder than OD and that
 tailored SD methods need to be developed for addressing effectively this significant problem.
}







\section{Introduction}



The detection of object states in images is  a problem of both theoretical and practical importance. By object state we refer to a condition of that object at a particular moment in time.  Some object states are mutually exclusive (e.g., open/closed), while others may hold simultaneously (e.g., open, filled, lifted). The transition from one state to another is, typically, the result of an action being performed upon the object. The state(s) in which an object can be found determine(s) to a large degree its behavior in the context of its interaction with other objects and entities.

Apart from  being a challenging task  bearing some unique characteristics, object state detection (SD) is a key visual competence, as the successful interaction of an agent with its environment depends critically on its ability to solve this problem. SD is also closely related to other important computer vision and AI problems, such as action recognition and planning. Surprisingly, the amount of research on this subject remains low, especially when juxtaposed with the vast research effort that has been invested over the last years in related computer vision problems, such as object detection and image classification. 

%


\begin{figure}
\begin{tabular}{llll}
\includegraphics[width=.2\linewidth,height=.2\linewidth]{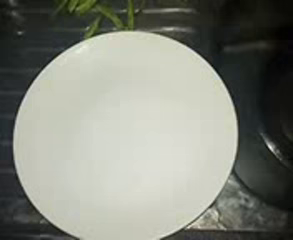} & \includegraphics[width=.2\linewidth,height=.2\linewidth]{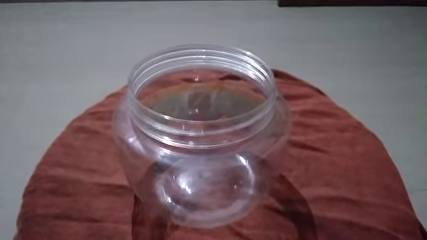} & \includegraphics[width=.2\linewidth,height=.2\linewidth]{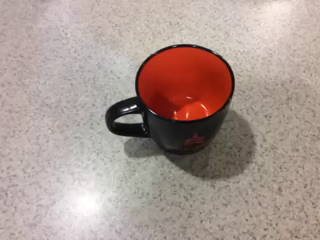}
 
 & \includegraphics[width=.2\linewidth,height=.2\linewidth]{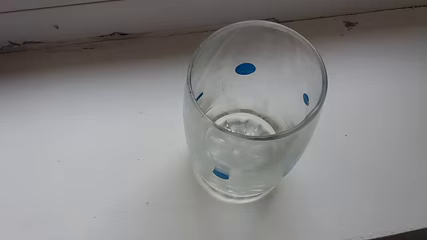}\\
 
\includegraphics[width=.2\linewidth,height=.2\linewidth]{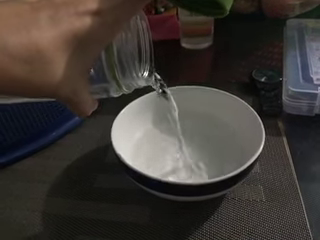} & \includegraphics[width=.2\linewidth,height=.2\linewidth]{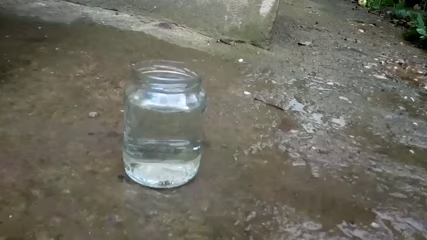}& \includegraphics[width=.2\linewidth,height=.2\linewidth]{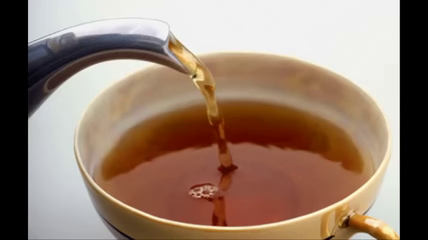}
 
 & \includegraphics[width=.2\linewidth,height=.2\linewidth]{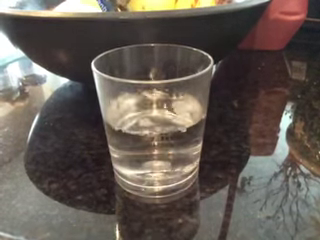}\\
 
\includegraphics[width=.2\linewidth,height=.2\linewidth]{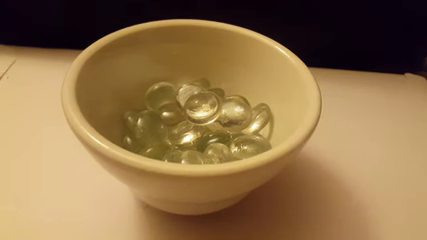} & \includegraphics[width=.2\linewidth,height=.2\linewidth]{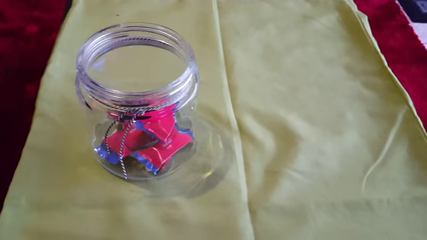} & \includegraphics[width=.2\linewidth,height=.2\linewidth]{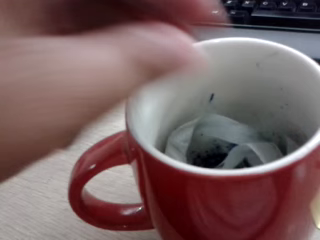}
 
 & \includegraphics[width=.2\linewidth,height=.2\linewidth]{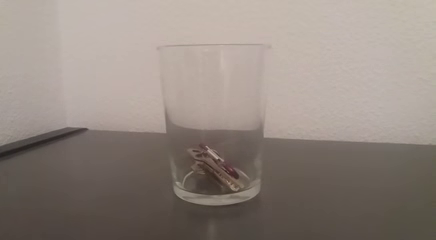}\\ 

\end{tabular}
\caption{Each row contains objects of the same state and each column contains objects of the same class. The variance of visual appearance is significantly greater for objects of the same state than for objects of the same category.}
\label{imag:ex1}
 \end{figure}

There are several arguments that attest the significance of a solution to the SD problem. First, the  detection of states is critical for decision making. In dynamic worlds, the conditions for stopping an action is state-dependent. For example, in order for a glass to be filled without an overflow its \textit{filled} state must be recognized or even predicted on time. Actually, in many cases, miss-classifying the object state could be equally or even more detrimental as miss-classifying its category (e.g., recognizing a bottle as a jar vs. recognizing a filled bottle as an empty bottle). Second, inferring correctly the states of  objects could facilitate significantly the recognition of actions. Conversely, the recognition of  an action can provide  cues about the states of objects which  were affected by the action. 
SD is also relevant to another active and important research area, that of object affordances recognition. The inference of object affordances depends critically on their current state. For example, many different kinds of objects\footnote{bottles, pots, plates, cups, vases amongst others.}  could be used for carrying liquids, provided that they are empty. In this case, inferring the state of the candidate container is more important that recognizing its class. 

\begin{figure}[t]
\includegraphics[width=\linewidth]{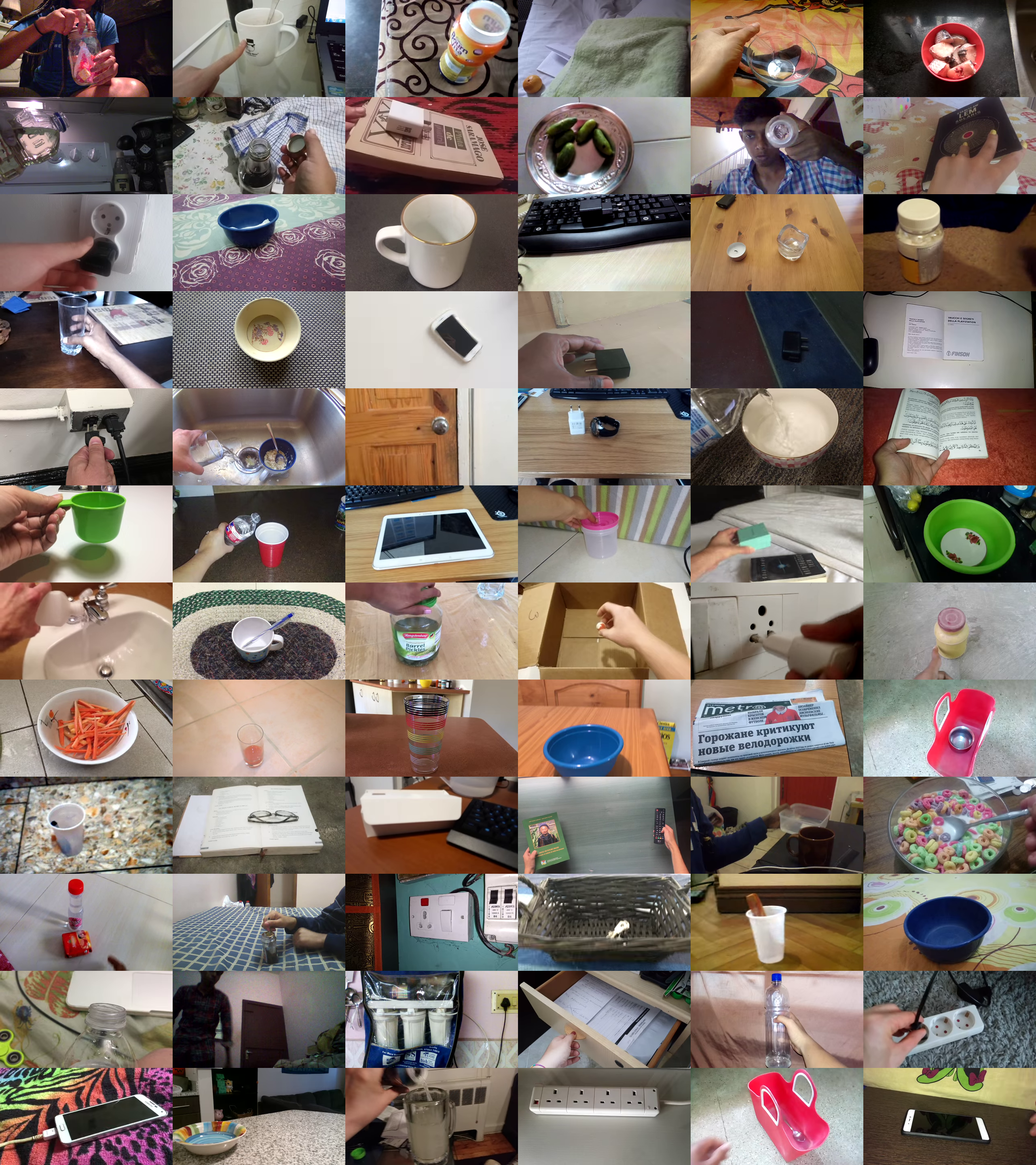} 
\caption{A small sample of the proposed OSDD dataset.}
\end{figure}

At a first glance, SD appears to be just a special case of object detection (OD). As an example, one could specialize a ``box'' detector to come up with an ``open box'' detector. This idea could explain the low levels of research activity  devoted to SD, per se. However, such an approach lacks scalability as the space of all object categories times the number of their possible states is huge. Moreover, there exist some important differences between the SD and OD problems. First, the intra-class variation for object states is vastly greater than the one for objects classes. For example, objects that are visually very dissimilar such as books, bottles,  boxes and  drawers, may belong to the same state class (open). Figure~\ref{imag:ex1} presents analogous examples. Moreover, the inter-class boundaries for the SD problem may hinge on minute details. For example, a slightly lifted  cap is the only difference between an open and a closed bottle. Another special aspect of SD has to do with the fact that an object may possess, simultaneously,   several non mutually exclusive states (e.g., an open, filled glass). Furthermore, considering the  problem of SD under a broader perspective which also includes videos as input, we can attest the dynamic nature of object states. For example, a ``filled'' cup may become ``empty'' in seconds in contrast to its object category which remains fixed.

Motivated by the previous observations and arguments which point towards the significance and the special characteristics of SD, as well as the limited study upon the subject, we investigate the SD problem in more detail. First, we provide a new  dataset, the \textit{Object States Detection Dataset (OSDD)}, consisting of  everyday household objects appearing in a variety of different states. Given the small number of related datasets, we believe that OSDD could be useful to anyone interested in the problem of SD. Second, we conduct a number of carefully devised experiments in order to  examine the performance of OD and SD on OSDD. The  experimental evaluation exposes the performance of solutions to the OD and SD problems, confirming that SD is harder than OD and attesting the need for robust and performant solutions to the SD problem.

\begin{table}[t]
    \centering
       \resizebox{0.92\textwidth}{!}{\begin{minipage}{\textwidth}
    \begin{tabular}{|l|c|c|c|c|c|c|}
    \hline
  \diagbox[innerleftsep=.5cm,innerrightsep=5.5pt]{{\bf Object}}{{\bf Pair}}    & \textbf{P1} & \textbf{P2} &  \textbf{P3}  &  \textbf{P4}  &  \textbf{P5} \\ \hline \hline
       
      Bottle, Jar, Tub  & \multirow{1}{*}{ \cmark} & \multirow{1}{*}{ \cmark} & \multirow{1}{*}{ \cmark} & \multirow{1}{*}{ \xmark} & \multirow{1}{*}{ \xmark} \\
\hline  

      Book, Drawer, Door & \multirow{1}{*}{ \cmark} & \multirow{1}{*}{ \xmark} & \multirow{1}{*}{ \xmark} & \multirow{1}{*}{ \xmark} & \multirow{1}{*}{ \xmark} \\
      	\hline 

  	   Basket, Box & \multirow{1}{*}{ \cmark} & \multirow{1}{*}{ \xmark} & \multirow{1}{*}{ \cmark} & \multirow{1}{*}{ \xmark} & \multirow{1}{*}{ \xmark} \\
	\hline
    
        Cup, Mug, Glass, Bowl & \multirow{1}{*}{ \xmark} & \multirow{1}{*}{ \cmark} & \multirow{1}{*}{ \cmark} & \multirow{1}{*}{ \xmark} & \multirow{1}{*}{ \xmark} \\
 \hline

    Phone, Charger, Socket & \multirow{1}{*}{ \xmark} & \multirow{1}{*}{ \xmark} & \multirow{1}{*}{ \xmark} & \multirow{1}{*}{ \cmark} & \multirow{1}{*}{ \xmark} \\
\hline

    Towel, Shirt, Newspaper  & \multirow{1}{*}{ \xmark} & \multirow{1}{*}{ \xmark} & \multirow{1}{*}{ \xmark} & \multirow{1}{*}{ \xmark} & \multirow{1}{*}{ \cmark} \\
\hline     
      
    \end{tabular}
    
    \end{minipage}}
    \caption{Objects and states in OSDD. Rows correspond clusters of object categories that may appear in the same set of states. Columns correspond to five pairs (P1-P5) of mutually exclusive states. P1:~Open/Close, P2:~Empty/Containing Liquid (CL), P3: Empty/Containing Solid (CS), P4:~Plugged/Unplugged, P5:~Folded/Unfolded.}
    \label{tab:dataset1}
\end{table}

\begin{table*}[thbp]
\centering

    \resizebox{0.88\textwidth}{!}{\begin{minipage}{\textwidth}

    \begin{tabular}{|l|c|c|c|c|c|c|c|c|c||r|}
    \cline{1-11}
    &  \rotatebox[origin=c]{0}{\textbf{Open}}    &   \rotatebox[origin=c]{0}{\textbf{Closed}}    &  \rotatebox[origin=c]{0}{\textbf{Empty}}   &  \rotatebox[origin=c]{0}{\textbf{CL}}  &  \rotatebox[origin=c]{0}{\textbf{CS}}   &  \rotatebox[origin=c]{0}{\textbf{Plugged}}  &  \rotatebox[origin=c]{0}{\textbf{Unplugged}}  &  \rotatebox[origin=c]{0}{\textbf{Folded}}  &  \rotatebox[origin=c]{0}{\textbf{Unfolded}}       &   \rotatebox[origin=c]{0}{\textbf{Total}}     \\ \hline \hline

basket &\cellcolor[HTML]{656565} &\cellcolor[HTML]{656565}&122&\cellcolor[HTML]{656565}  &  336&\cellcolor[HTML]{656565}  &  \cellcolor[HTML]{656565}  &  \cellcolor[HTML]{656565}  &  \cellcolor[HTML]{656565}  &  458 \\ \hline
book &316&679&\cellcolor[HTML]{656565}  &  \cellcolor[HTML]{656565}  &  \cellcolor[HTML]{656565}  &  \cellcolor[HTML]{656565}  &  \cellcolor[HTML]{656565}  &  \cellcolor[HTML]{656565}  &  \cellcolor[HTML]{656565}  &  995 \\ \hline
bottle &891&923&420&803&238&\cellcolor[HTML]{656565}  &  \cellcolor[HTML]{656565}  &  \cellcolor[HTML]{656565}  &  \cellcolor[HTML]{656565}  &  3275 \\ \hline
bowl &\cellcolor[HTML]{656565}  &  \cellcolor[HTML]{656565}  &  809&146&790&\cellcolor[HTML]{656565}  &  \cellcolor[HTML]{656565}  &  \cellcolor[HTML]{656565}  &  \cellcolor[HTML]{656565}  &  1745 \\ \hline
box &518&291&184&\cellcolor[HTML]{656565}  &  337&\cellcolor[HTML]{656565}  &  \cellcolor[HTML]{656565}  &  \cellcolor[HTML]{656565}  &  \cellcolor[HTML]{656565}  &  1330 \\ \hline
charger &\cellcolor[HTML]{656565}  &  \cellcolor[HTML]{656565}  &  \cellcolor[HTML]{656565}  &  \cellcolor[HTML]{656565}  &  \cellcolor[HTML]{656565}  &  235&376&\cellcolor[HTML]{656565}  &  \cellcolor[HTML]{656565}  &  611 \\ \hline
cup &\cellcolor[HTML]{656565}  &  \cellcolor[HTML]{656565}  &  432&139&220&\cellcolor[HTML]{656565}  &  \cellcolor[HTML]{656565}  &  \cellcolor[HTML]{656565}  &  \cellcolor[HTML]{656565}  &  791 \\ \hline
door &271&481&\cellcolor[HTML]{656565}  &  \cellcolor[HTML]{656565}  &  \cellcolor[HTML]{656565}  &  \cellcolor[HTML]{656565}  &  \cellcolor[HTML]{656565}  &  \cellcolor[HTML]{656565}  &  \cellcolor[HTML]{656565}  &  752 \\ \hline
drawer &468&484&\cellcolor[HTML]{656565}  &  \cellcolor[HTML]{656565}  &  \cellcolor[HTML]{656565}  &  \cellcolor[HTML]{656565}  &  \cellcolor[HTML]{656565}  &  \cellcolor[HTML]{656565}  &  \cellcolor[HTML]{656565}  &  952 \\ \hline
glass &\cellcolor[HTML]{656565}  &  \cellcolor[HTML]{656565}  &  523&363&215&\cellcolor[HTML]{656565}  &  \cellcolor[HTML]{656565}  &  \cellcolor[HTML]{656565}  &  \cellcolor[HTML]{656565}  &  1101 \\ \hline
jar &356&295&176&111&369&\cellcolor[HTML]{656565}  &  \cellcolor[HTML]{656565}  &  \cellcolor[HTML]{656565}  &  \cellcolor[HTML]{656565}  &  1307 \\ \hline
mug &\cellcolor[HTML]{656565}  &  \cellcolor[HTML]{656565}  &  541&160&269&\cellcolor[HTML]{656565}  &  \cellcolor[HTML]{656565}  &  \cellcolor[HTML]{656565}  &  \cellcolor[HTML]{656565}  &  970 \\ \hline
newspaper &\cellcolor[HTML]{656565}  &  \cellcolor[HTML]{656565}  &  \cellcolor[HTML]{656565}  &  \cellcolor[HTML]{656565}  &  \cellcolor[HTML]{656565}  &  \cellcolor[HTML]{656565}  &  \cellcolor[HTML]{656565}  &  322&135&457 \\ \hline
phone &\cellcolor[HTML]{656565}  &  \cellcolor[HTML]{656565}  &  \cellcolor[HTML]{656565}  &  \cellcolor[HTML]{656565}  &  \cellcolor[HTML]{656565}  &  205&743&\cellcolor[HTML]{656565}  &  \cellcolor[HTML]{656565}  &  948 \\ \hline
shirt &\cellcolor[HTML]{656565}  &  \cellcolor[HTML]{656565}  &  \cellcolor[HTML]{656565}  &  \cellcolor[HTML]{656565}  &  \cellcolor[HTML]{656565}  &  \cellcolor[HTML]{656565}  &  \cellcolor[HTML]{656565}  &  139&187&326 \\ \hline
socket &\cellcolor[HTML]{656565}  &  \cellcolor[HTML]{656565}  &  \cellcolor[HTML]{656565}  &  \cellcolor[HTML]{656565}  &  \cellcolor[HTML]{656565}  &  486&1016&\cellcolor[HTML]{656565}  &  \cellcolor[HTML]{656565}  &  1502 \\ \hline
towel &\cellcolor[HTML]{656565}  &  \cellcolor[HTML]{656565}  &  \cellcolor[HTML]{656565}  &  \cellcolor[HTML]{656565}  &  \cellcolor[HTML]{656565}  &  \cellcolor[HTML]{656565}  &  \cellcolor[HTML]{656565}  &  320&197&517 \\ \hline
tub &276&153&136&\cellcolor[HTML]{656565}&416&\cellcolor[HTML]{656565}  &  \cellcolor[HTML]{656565}  &  \cellcolor[HTML]{656565}  &  \cellcolor[HTML]{656565}  &  981 \\ \hline \hline
Total &3096&3306&3343&1722&3190&926&2135&781&519&19018 \\ \hline

    \end{tabular}
\end{minipage}}
\begin{center}
    \caption{Number of annotations for the different object-state combinations.}
    \label{tab:dataset2}
\end{center}
\end{table*}

\begin{table*}[t]

    \centering
    \begin{tabular}{|l|l|r|r|r|l|l|l|}
    \hline
   \textbf{Dataset}   & \textbf{Images/Videos} & \textbf{Annotations} & \textbf{States}  & \textbf{Objects} & \textbf{Task} &  \textbf{View}  \\ \hline\hline

        OSDD  (ours, proposed) & 13,744 images &  19,018  &9 & 18 & SD & 3rd person \\ \hline   
      \cite{Isola2015}   & 63,440 images   & 63,440  & 18 & NA &  SC & Egocentric \\ \hline
       \cite{Liu2017}   & 809 videos  & 330,000  & 21 & 25 & SD \& AR & Egocentric \\ \hline
         \cite{fire2017inferring}   & 490 Videos  & 180,374
 & 17 & 13 & SD \& AR & 3rd person \\ \hline
       \cite{Alayrac17}   & 630 Videos  & 19,949  & 7 & 5 & SD \& AR & 3rd person  \\ \hline
     \end{tabular}
    \caption{OSDD and existing object states datasets, in numbers.}
    \label{tab:my_label}
\end{table*}

\section{Related Work}
We review existing approaches to the OD and SD problems. For OD the literature is vast and its detailed review is beyond the scope of this paper. Therefore, we restrict ourselves to an overview of the main classes of approaches that follow the more recent,  state-of-art deep-learning paradigm. We also provide pointers to existing object states datasets and we identify the contributions of this work.   
%

%

\vspace*{0.2cm}\noindent\textbf{Object Detection:} Deep learning-based object detection frameworks can be categorized into two groups: (i)~one-stage detectors, such as YOLO~\cite{redmon2017yolo9000} and SSD~\cite{liu2016ssd} and (ii)~two-stage detectors, such as Region-based CNN (R-CNN)~\cite{girshick2015fast}  and Pyramid Networks~\cite{lin2017feature}. Two-stage detectors  use a proposal generator to create a sparse set of proposals in order to extract features from each proposal which are followed by region classifiers that make predictions about the category of the proposed region, whereas one-stage detectors generate  categorical predictions of objects on each location of the feature maps omitting the cascaded region classification procedure. Two-stage detectors are typically more performant and achieve state-of-the-art results on the majority of the public benchmarks, while one-stage detectors are characterized by computational efficiency and are used more widely for real-time OD.  Some other highly influential works  in the field are~\cite{he2016deep}, \cite{huang2017densely}, \cite{he2017mask} and \cite{lin2017focal}.

\vspace*{0.2cm}\noindent\textbf{State Detection:} 
The research that has been conducted in SD has treated the problem using either videos or images as input. In the first case, the problem of SD usually serves as a stepping stone to achieve action recognition.

In \cite {Alayrac17},  SD is studied in the context of videos containing manipulation actions performed upon 7 classes of objects. The authors formulate SD as a discriminative clustering problem and attempt to address it by optimization methods. \cite{Liu2017} represent state-altering actions as concurrent and sequential object fluents (states) and utilize a beam search algorithm  for  fluent detection and action recognition.  In a similar vein, \cite{Aboubakr2019} explores state detection in tandem with action recognition. The method is based on the learning of appearance models of objects and their states from video frames which are used in conjunction with a state transition matrix which maps action labels into a pre-state and a post-state. 
In \cite{Isola2015}, the states and transformations of objects/scenes on image collections are studied, and the learned state representations are extended to different object classes. \cite{Fire2015,fire2017inferring} examines the causal relations between human actions and object fluent changes. \cite{Fathi2013} developed a weakly supervised method to recognize actions and states of manipulated objects before and after the action proposing a weakly supervised method for learning the object and material states that are necessary for recognizing daily actions. 
\cite{wang2016actions}  designed a Siamese network to model precondition states, effect states and their associate actions.  \cite{Bertasius2020} leveraged the semantic and compositional structure of language by training a visual detector to predict a contextualized word embedding of the object and its associated narration enabling object representation learning  where concepts relate to a semantic language metric.

\begin{table}[t]

    \centering
    \resizebox{0.90\textwidth}{!}{\begin{minipage}{\textwidth}
    \begin{tabular}{|l|r|l|r|r|l|}
    \hline
 \rotatebox[origin=c]{35}{\textbf{Scenario}}    &
  \rotatebox[origin=c]{35}{\textbf{Obj.}}    &
     \rotatebox[origin=c]{35}{\textbf{States}}    &
      \rotatebox[origin=c]{35}{\textbf{Nets}}    &
      \rotatebox[origin=c]{35}{\textbf{Exp.}}    &
       \rotatebox[origin=c]{35}{\textbf{Tasks}}   

     \\   \hline\hline

OOOS& 1 & 2 ME & 28  & 28  & SD     \\   \hline
MOOS& 3-6 &2 ME & 5 & 5  & SD  \\ \hline
OOMS& 1 & 3-5 &  8  &  8 & SD   \\ \hline
MOMS& 18 & 9  & 1 & 1   & SD  \\ \hline
ODS& 18 & 9   & 1  & 1  & OD  \\ \hline
TOOS& 2 & 2 ME   & 28  & 7 & SD, OD  \\ \hline
SDG& 3-9 & 2 ME  & 0  & 5  & SD  \\ \hline     

    \end{tabular}
    
    \end{minipage}}
    \caption{The experimental details of the different scenarios. Columns show the scenario name, number of involved objects, number of involved states (ME: mutually exclusive), number of trained networks, number of performed experiments, target problem.}
        \label{tab:exprs}
\end{table}

Overall, previous works dealt with the SD problem mostly by employing egocentric videos. These are less challenging than 3rd person-view input due to the higher resolution in which objects are imaged and the lack of considerable clutter/occlusions. Additionally, in the majority of the cases, SD is considered a means for achieving AR and not as a goal on its own right, thus the  studies of the SD problem are of a limited scope. 


%
%


\vspace*{0.2cm}\noindent\textbf{Object States Datasets:} Currently, there are only a few object states datasets available to the research community. The MIT-States dataset~\cite{Isola2015} is composed of 63,440 images involving 115 attribute classes and 245 object categories and is suitable for state classification (SC). However, the dataset contains segmented images of objects in white background so they are far from being natural. Additionally, most of the 115 reported attributes constitute object properties that are not necessarily object states. 
\cite{Liu2017} presents a dataset consisting of 809 videos covering 21 state classes and 25 object categories. The scope of the  dataset is SD and action recognition (AR). 
The dataset presented in \cite{Alayrac17} contains 630 videos concerning 7 state classes and 5 object categories, also suited for SD and AR. Finally, \cite{fire2017inferring} proposes a  RGBD video dataset for causal reasoning  spanning  13 object and 
17 state categories.

In general, the existing SD datasets  lack the diversity that characterizes the OD datasets, which can be attributed to the fact that they have been created without having the task of SD  as a primary goal. 

\vspace*{0.2cm}\noindent\textbf{Our contribution:} We introduce OSDD, a new states dataset of 13,744 images and 19,018 states annotations. 
The dataset involves 18 object categories that may appear in 9 different state classes.  The images of the dataset are characterised  by a great variety regarding viewing angles, background and foreground scene settings, object sizes and object-state combinations. These characteristics render the dataset more challenging than most of the existing datasets which contain a much more limited diversity with respect to the aforementioned characteristics. Moreover, OSDD could be proven useful for those who want to assess the performance of approaches addressing the tasks of zero-shot and few-shot recognition.

\begin{table}[t]
	\footnotesize
    \centering
     \resizebox{0.90\textwidth}{!}{\begin{minipage}{\textwidth} 
    \begin{tabular}{|l|r|r|r|r|}
    
   \hline 
      
 \textbf{Metric } & \textbf{OOOS}  & \textbf{MOOS}   & \textbf{OOMS}     &   \textbf{MOMS}   \\ \hline \hline
 
  Weighted AP &  \color{blue}68.4   & 63.2 &  54.3 &  \color{red}46.6 \\   \hline 
  
   Best perf. (\%) &  \color{blue}58.0   & 38.0 &  \color{red}2.0 &  \color{red}2.0 \\   \hline 
  Worst perf. (\%) &  \color{blue}6.0 &   \color{blue}6.0 &  14.0 &  \color{red}74.0 \\   \hline   \hline 
    Avg  mAP &  \color{blue}62.3 &  56.5.0  & 46.8   &  \color{red}38.4 \\    \hline 
     Avg  AP@50:5:95 & \color{blue} 40.5 &     39.7   &  29.1  &\color{red}23.7 \\ 
    \hline

  \end{tabular}
  
  \end{minipage}}
   \caption{Aggregate results for scenarios OOOS, MOOS, OOMS and MOMS. Row 1: weighted  AP, row 2: percentage of experiments in which this scenario achieved best AP, row 3: percentage of experiments in which this scenario achieved worst AP, row 4: Average mAP , row 5: Average AP@50:5:95. Blue/red font indicates best/worst performance in a line.}
   \label{tab:table_aggr}
\end{table}

In addition, we provide an extensive quantitative, experimental evaluation of several aspects of the SD problem. We conduct 55 different experiments  including 71 differently trained networks in the context of 7 scenarios of varied settings which allows us to draw valuable insights about the nature of SD and its relation to OD. OSDD and the associated experiments set useful baselines for assessing progress on the topic of SD. To the best of our knowledge, our work is the first in which such a thorough investigation of the SD task is performed.

\section{The Object States Detection Dataset (OSDD)}
The proposed  Objects States Detection Dataset (OSDD)\footnote{\url{https://github.com/philipposg/OSDD}} consists of images depicting everyday household objects in a number of different states. The ground-truth annotations  involve the labels and bounding boxes spanning 18 object categories and 9 state classes. 
The object categories are: \textit{bottle, jar, tub, book, drawer, door, cup, mug, glass, bowl, basket, box, phone, charger, socket, towel, shirt} and \textit{newspaper}. The 9 state classes are: \textit{open, close, empty, containing something liquid (CL), containing something solid (CS), plugged, unplugged, folded} and \textit{unfolded}.

The states are grouped in 5 pairs of mutually exclusive states. Table~\ref{tab:dataset1} shows which cluster of objects is relevant to which pair of mutually exclusive (ME) states.  Table~\ref{tab:dataset2} provides an overview of the dataset contents. For all object categories (rows) and states (columns), we report the number of annotations.


\begin{table}[t]

    \centering
\resizebox{0.85\textwidth}{!}{\begin{minipage}{\textwidth}     
  \begin{tabular}{|l|r|r||r|r|r|}
    \hline  
     \textbf{State}  &  \textbf{OOOS}  & \textbf{MOOS}   & \textbf{OOMS}     &   \textbf{MOMS}   \\ \hline   \hline
     
\textbf{Open} & \color{blue} 68.7 & 67.4 & 52.4 &  \color{red}49.9 \\ \hline
\textbf{Closed} &  \color{blue}69.2 & 62.2 & 62.4 & \color{red}48.2 \\ \hline
\textbf{Empty(vs CL)} &  \color{blue}70.0 & 67.6 & 58.3 &\color{red} 56.4 \\ \hline
\textbf{Filled} & \color{blue} 49.9 & 40.4 & 38.4 & \color{red}25.0 \\ \hline
\textbf{Empty(vs CS)} &  \color{blue}73.0 & 69.8 & 58.3 &\color{red} 56.4 \\ \hline
\textbf{Occupied} &  \color{blue}64.8 & 56.0 & 53.6 & \color{red}37.2 \\ \hline
\textbf{Folded} & 57.3 &  \color{blue}62.0 & NA &\color{red} 39.5 \\ \hline
\textbf{Unfolded} & 37.0 &  \color{blue}39.9 & NA & \color{red}22.7 \\ \hline
\textbf{Connected} &  \color{blue}60.3 & 49.2 & NA & \color{red}24.8 \\ \hline
\textbf{Unconnected} &  \color{blue}84.9 & 78.5 & NA &\color{red} 66.1 \\ \hline

 \end{tabular}
   \end{minipage}}
 \caption{ Experimental results for the AP metric averaged at the level of states for the OOOS, MOOS, OOMS and MOMS scenarios.  Each row corresponds to a different state.
 Blue/red font indicates best/worst performance for the given state.}
    \label{table:all3}
\end{table}

The images were obtained by selecting video frames   from the something-something V2 Dataset~\cite{mahdisoltani2018effectiveness}.
Specifically, images containing visually salient objects and states of the aforementioned categories were captured and annotated with bounding-boxes and ground truth labels referring to the corresponding object categories and state classes. Overall, the dataset contains  13,744 images and 19,018 annotations obtained by selecting the first, last and middle frames of 9,015 videos, after checking that each of them contains salient information.
There are more annotations than images because (a) in a certain image there may be more than one objects or/and (b) a certain object is annotated for all the non-exclusive states it appears in.

The dataset annotation was performed based on the Computer Vision Annotation Tool (CVAT)\footnote{{https://github.com/openvinotoolkit/cvat}}. In order to handle properly ambiguous situations and safeguard from erroneous annotations, each image was examined at least five times. Overall, the annotation process required approximately 350 person hours.

\begin{table}[t]

    \centering
    
\footnotesize   
  \begin{tabular}{|l|r|r||r|r|r|}
    \hline  
     \textbf{Object}  &  \textbf{OOOS}  & \textbf{MOOS}   & \textbf{OOMS}     &   \textbf{MOMS}   \\ \hline   \hline
     
\textbf{bottle} & \color{blue}61.9 & 50.0 & 40.2 &  \color{red}32.4\\ \hline
\textbf{tub} & \color{blue}74.8 & 48.1 & 47.0 &  \color{red}18.9\\ \hline
\textbf{cup} & 70.8 & \color{blue}73.2 & 66.5 &  \color{red}64.7\\ \hline
\textbf{mug} & 78.0 & \color{blue}80.3 & 82.6 &  \color{red}71.4\\ \hline
\textbf{jar} & 50.8 & \color{blue}59.3 & 36.0 & \color{red} 30.6 \\ \hline
\textbf{glass} &\color{blue} 64.2 & 62.6 & 55.9 &  \color{red}50.0 \\ \hline

\textbf{bowl} & 75.5 & \color{blue}77.3 & 71.5 &  \color{red}68.0 \\ \hline

\textbf{box} &\color{blue} 73.4 & 71.2 &  \color{red}59.6 & 62.9\\ \hline
\textbf{basket} & 58.6 & \color{blue}65.6 & NA &  \color{red}55.2 \\ \hline
\textbf{book} & \color{blue}74.7 & 63.9 & NA &  \color{red}62.0 \\ \hline
\textbf{door} & \color{blue}45.9 & 45.4 & NA &  \color{red}30.7 \\ \hline
\textbf{shirt} & \color{blue}40.4 & 33.3 & NA & \color{red} 22.8 \\ \hline
\textbf{newspaper} & 40.5 & \color{blue}45.8 & NA &  \color{red}14.5 \\ \hline
\textbf{towel} & 54.7 & \color{blue}62.4 & NA &  \color{red}44.1 \\ \hline
\textbf{drawer} & \color{blue}73.8 & 66.5 & NA &  \color{red}58.5 \\ \hline
\textbf{phone} & \color{blue}79.2 & \color{blue}79.2 & NA &  \color{red}70.0 \\ \hline
\textbf{charger} &\color{blue} 69.5 & 38.7 & NA &  \color{red}35.2 \\ \hline
\textbf{socket} & \color{blue}80.3 & 75.8 & NA &  \color{red}51.9 \\ \hline

 \end{tabular}
 \caption{ Experimental Results for the AP metric averaged at the level of states for the OOOS, MOOS, OOMS and MOMS scenarios. Each row corresponds to a different object.
 Blue/red font indicates best/worst performance for the given object.}
    \label{table:all4}
\end{table}

\subsection{OSDD vs existing object states datasets}
Table~\ref{tab:my_label} summarizes information regarding the existing SD datasets. The dataset that is most similar to OSDD is the one presented in \cite{Isola2015}. However, the two datasets differ in a number of ways. First, the images in our dataset are snapshots of video tracklets that show objects manipulations, whereas the images in \cite{Isola2015} stem from scrapping. 
Moreover, the majority of images in \cite{Isola2015} are cropped containing a single object, whereas OSDD images contain objects in context (e.g., other objects,  background, etc). 

Another important remark regarding the existing datasets, is that in the case  of~\cite{Liu2017},~\cite{fire2017inferring}  and~\cite{Alayrac17}, annotations were provided for adjacent video frames, whereas in our case they involve the first, middle and last frames of 9,015 videos. Thus, although they appear to provide many more annotations, they are much less diverse than OSDD due to the similarity between many of the annotated frames.

In summary, the aspects that  distinguish OSDD from the existing datasets are its  greater diversity regarding objects and states appearance, the greater background/foreground variation and the greater diversity of viewpoints which results in a vast variety of object sizes and viewing angles. These characteristics  make it unique for studying SD in challenging, realistic  scenarios.

\begin{table*}[h]

    \centering
    
\footnotesize

    \begin{tabular}{|l|l|r|r||c|c|r|}
    \hline  
    \textbf{Object} & \textbf{Pair}  &  \textbf{OOOS}  & \textbf{MOOS}   & \textbf{OOMS}     &   \textbf{MOMS}   \\ \hline   \hline
     
\multirow{3}{*}{\textbf{bottle}} & \textbf{P1} & \color{blue}71.5   \slash \ 48.0 &   \color{red}67.7  \slash \ 46.8 &   \multirow{3}{*}{40.4 \slash  \  25.3 }  & \multirow{31}{*}{38.4   \slash \ 23.7}\\ \cline{2-4}
     & \textbf{P2} &  \color{blue}38.0  \slash  \  26.3 & \color{red}24.4 \slash  \ 15.1   & & \\  \cline{2-4}
         & \textbf{P3} & \color{blue} 68.6  \slash  \ 43.8  &  \color{red}43.8  \slash  \ 30.1 & & \\ \cline{1-5}
         
 \multirow{3}{*}{\textbf{jar}}
  & \textbf{P1}&  \color{red}50.0 \slash \  25.8  &  \color{blue}67.2   \slash \ 49.8 & \multirow{3}{*}{ 33.9 \slash \  22.3} & \\  \cline{2-4}
  & \textbf{P2} &\color{red}35.8 \slash \ 25.0   & \color{blue} 52.2   \slash \ 33.7 & &  \\  \cline{2-4}
  & \textbf{P3}&   \color{red}51.8   \slash \  33.4 &  \color{blue}52.6   \slash \  38.9&  & \\ \cline{1-5}

    \multirow{3}{*}{\textbf{tub}} & 
    \textbf{P1} & \color{blue}83.0  \slash \ 59.1 &   \color{red}57.5  \slash \ 42.4  &  \multirow{3}{*}{41.9 \slash \ 23.2} &\\  \cline{2-4}
  & \textbf{P2} & \color{red}44.5  \slash \  25.3 &  \color{blue} 66.6   \slash \ 37.5 & & \\ \cline{2-4}
  & \textbf{P3} &  \color{blue}64.2 \slash \  39.7 &  \color{red}29.4  \slash \  20.6& & \\ \cline{1-5}

  \multirow{2}{*}{\textbf{box}}
  & \textbf{P1} & \color{blue}   73.0  \slash \  53.4 &  \color{red} 61.1  \slash \ 46.2 &\multirow{2}{*}{ 56.7  \slash \  20.7} & \\  \cline{2-4}
  & \textbf{P3} &  \color{blue} 67.2  \slash \  43.8 &  \color{red}60.4 \slash \   45.0 & & \\  \cline{1-5}

  \multirow{2}{*}{\textbf{glass}}
  & \textbf{P2} & \color{blue}60.1 \slash \  41.5 &   \color{red}52.4 \slash \  34.9 & \multirow{2}{*}{46.8 \slash \ 32.9 } &  \\ \cline{2-4}
  & \textbf{P3} &  \color{red}57.9 \slash \  41.0 &  \color{blue}63.7 \slash \  47.7 &  & \\ \cline{1-5}

\multirow{2}{*}{\textbf{cup}}
  & \textbf{P2} & \color{blue}59.9 \slash \  41.6 &   \color{red}54.5 \slash \  36.6 & \multirow{2}{*}{53.5 \slash \ 36.5 } &   \\ \cline{2-4}
  & \textbf{P3} &  \color{red}62.2 \slash \  47.8 &  \color{blue}75.0 \slash \  60.0 & & \\ \cline{1-5}

\multirow{2}{*}{\textbf{mug}}
  & \textbf{P2} &  \color{red}56.1 \slash \  36.6 &  \color{blue}56.4 \slash \  36.9 & \multirow{2}{*}{ 64.8 \slash \ 50.1} & \\   \cline{2-4}
  & \textbf{P3} &  \color{red}64.6 \slash \  45.8 &  \color{blue}73.7 \slash \  56.3 & & \\ \cline{1-5}

 \multirow{2}{*}{\textbf{bowl}}
   & \textbf{P2} &  \color{red}57.0 \slash \  40.6 &  \color{blue}78.7 \slash \  64.5 &\multirow{3}{*}{54.9 \slash \ 35.5}& \\ \cline{2-4}
  & \textbf{P3} & \color{blue}74.9 \slash \  52.4 &   \color{red}59.8 \slash \  41.8 &  & \\   \cline{1-5}

\multirow{1}{*}{\textbf{basket}}
  & \textbf{P3} &\color{blue} 55.6 \slash \  29.1 &   \color{red}53.6 \slash \  41.6 & NA & \\  \cline{1-5}

\textbf{phone}
  & \textbf{P4} &  \color{red}61.7 \slash \  40.6 &  \color{blue}64.2 \slash \  40.6 & NA & \\ \cline{1-5}

\textbf{charger}
  & \textbf{P4} & \color{blue}68.6 \slash \  42.2 &   \color{red}38.1 \slash \  21.4 & NA & \\ \cline{1-5}

\textbf{socket}
  & \textbf{P4} & \color{blue}76.2 \slash \  39.3 &   \color{red}71.0 \slash \  36.1 & NA & \\ \cline{1-5}

\textbf{book}
  & \textbf{P1} &\color{blue} 64.2 \slash \  44.3 &   \color{red}57.0 \slash \  42.7 & NA & \\ \cline{1-5}

\textbf{door} 
  & \textbf{P1} & \color{blue}42.1 \slash \  23.9 &   \color{red}39.1 \slash \  23.8 & NA & \\ \cline{1-5}

\textbf{drawer} 
  & \textbf{P1} & \color{blue}75.6 \slash \  43.9 &   \color{red}68.9 \slash \  45.8 & NA & \\ \cline{1-5}

\textbf{newspaper}
  & \textbf{P5} &  \color{red}35.7 \slash \  23.5 &  \color{blue}43.7 \slash \  29.8 & NA & \\ \cline{1-5}

\textbf{towel}
  & \textbf{P5} &  \color{red}53.9 \slash \  35.5 &  \color{blue}61.5 \slash \  41.6& NA & \\ \cline{1-5}

\textbf{shirt}
  & \textbf{P5} & \color{blue}41.8 \slash \  22.5 &   \color{red}36.5 \slash \  25.6& NA & \\ \hline \hline

\multicolumn{2}{|c|}{\textbf{Weighted  Average}}  &  \color{blue} 62.3   \slash \ 40.5 & 56.5  \slash \ 39.7   & 46.8  \slash \ 29.1   &   \color{red} 38.4  \slash \ 23.7   \\ \hline

 \end{tabular}
 \caption{ Experimental results  for the OOOS, MOOS, OOMS and MOMS scenarios. First  number in each cell corresponds to the mAP  metric and second number corresponds to the    AP@50:5:95 metric respectively. The metrics are calculated  at the level of  each object pairs for the OOOS and MOOS scenarios,   at the level of each object for the OOMS scenario and at the level of all objects for the MOMS scenario respectively.
 Blue/red font indicates best/worst performance between the OOOS and MOOS scenarios.}
    \label{table:all2}
\end{table*}

\begin{figure*}[t] 
    \centering
    \includegraphics[width=0.95\textwidth]{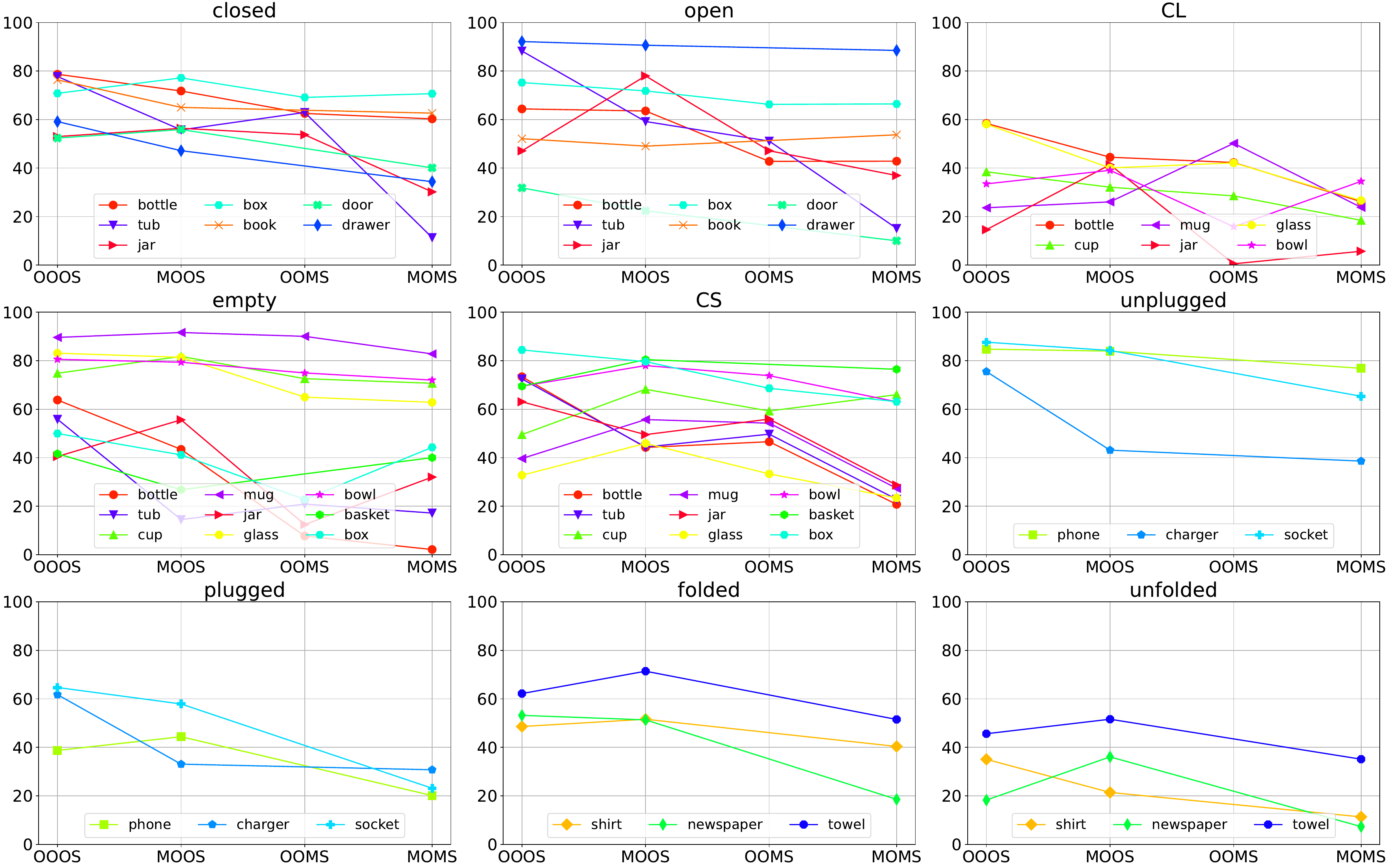}
   \caption{AP for the OOOS, MOOS, OOMS and MOMS scenarios, for each object state in OSDD.}
    \label{fig:exps}
   \end{figure*}

\begin{table}[t]
    \centering

    \begin{tabular}{|l|r|r|}
    \hline
        \textbf{Metric}& \textbf{MOMS} &  \textbf{ODS}    \\  \hline\hline

      \textbf{AP@50:5:95} & \color{red}23.7 &   \color{blue}30.9  \\   \hline
       \textbf{mAP}   &   \color{red}38.4    &  \color{blue}48.8   \\   \hline
      
\end{tabular}

   \caption{Experimental results for the MOMS and ODS scenarios. See text for details.}
    \label{tab:exprMOOS}
\end{table}

\begin{table*}[t]

     \resizebox{0.80\textwidth}{!}{\begin{minipage}{\textwidth} 
     
      \centering
    \begin{tabular}{|l|l||l|l||c|c|c|||c|c|c|}
     \hline
    
     \multicolumn{4}{|c||}{\textbf{Scenario Details}} &   \multicolumn{6}{|c|}{\textbf{Results}}    \\   \hline     
     
   \textbf{Object 1}&  \textbf{Object 2}&  \textbf{State 1}& \textbf{State 1}&
 \rotatebox[origin=c]{0}{\textbf{CS1 }}    &
  \rotatebox[origin=c]{0}{\textbf{CS 2  }}    &
       \rotatebox[origin=c]{0}{\textbf{ST }}    &

 \rotatebox[origin=c]{0}{\textbf{CO1 }}    &
  \rotatebox[origin=c]{0}{\textbf{CO2 } }    &  
  
      \rotatebox[origin=c]{0}{\textbf{OB}   }

     \\   \hline \hline 
  
   \textbf{bottle} &  \textbf{jar} &  \textbf{open} & \textbf{closed} &19.5 \slash  13.3  &\textcolor{red}{ 9.0} \slash \textcolor{red}{ 5.4}  & 27.2 \slash  17.6  & 26.7 \slash  18.1  &\textcolor{blue}{ 69.7} \slash \textcolor{blue}{ 45.5}  & 67.3 \slash \textcolor{blue}{ 45.5} \\ \hline 
   \textbf{bottle} &  \textbf{jar} &  \textbf{empty} & \textbf{CL} &7.7 \slash  5.0  &\textcolor{red}{ 1.5} \slash \textcolor{red}{ 0.7}  & 35.2 \slash  20.7  & 52.1 \slash  28.0  & 38.2 \slash  20.9  &\textcolor{blue}{ 58.8} \slash \textcolor{blue}{ 33.4} \\ \hline 
   \textbf{cup} &  \textbf{glass} &  \textbf{empty} & \textbf{CL} &2.4 \slash \textcolor{red}{ 1.6}  & 2.6 \slash  1.8  & 55.0 \slash  39.4  & 59.2 \slash  43.2  & 49.8 \slash  26.9  &\textcolor{blue}{ 79.8} \slash \textcolor{blue}{ 44.6} \\ \hline 
   \textbf{cup} &  \textbf{glass} &  \textbf{empty} & \textbf{CS} &4.4 \slash \textcolor{red}{ 3.0}  & 10.6 \slash  7.4  & 59.7 \slash  43.3  & 37.3 \slash  24.1  & 56.1 \slash  28.1  &\textcolor{blue}{ 75.4} \slash \textcolor{blue}{ 54.6} \\ \hline 
   \textbf{charger} &  \textbf{phone} &  \textbf{plugged} & \textbf{unplugged} &5.7 \slash  3.3  &\textcolor{red}{ 4.7} \slash \textcolor{red}{ 3.0}  & 49.4 \slash  28.3  & 49.9 \slash  30.8  & 67.0 \slash  33.8  &\textcolor{blue}{ 81.7} \slash \textcolor{blue}{ 45.9} \\ \hline 
   \textbf{shirt} &  \textbf{towel} &  \textbf{unfolded} & \textbf{folded} &19.7 \slash  14.4  &\textcolor{red}{ 13.0} \slash \textcolor{red}{ 7.2}  & 32.5 \slash  19.2  & 45.4 \slash  28.3  & 51.4 \slash  27.3  &\textcolor{blue}{ 60.3} \slash \textcolor{blue}{ 37.4} \\ \hline 
   \textbf{newspaper} &  \textbf{towel} &  \textbf{unfolded} & \textbf{folded} &0.3 \slash \textcolor{red}{ 0.2}  & 3.6 \slash  2.4  & 77.1 \slash \textcolor{blue}{ 58.6}  & 72.4 \slash  51.0  & 41.0 \slash  21.6  &\textcolor{blue}{ 84.4} \slash  54.4 \\ \hline 
   \textbf{bowl} &  \textbf{mug} &  \textbf{empty} & \textbf{CL} &5.2 \slash \textcolor{red}{ 3.5}  & 18.6 \slash  11.5  & 50.1 \slash  32.3  & 34.1 \slash  25.2  & 61.5 \slash  37.9  &\textcolor{blue}{ 83.9} \slash \textcolor{blue}{ 67.2} \\ \hline 
   \textbf{bowl} &  \textbf{mug} &  \textbf{empty} & \textbf{CS} &6.8 \slash \textcolor{red}{ 4.8}  & 21.5 \slash  14.0  & 46.0 \slash  35.0  & 14.3 \slash  9.2  & 67.6 \slash  39.6  &\textcolor{blue}{ 77.7} \slash \textcolor{blue}{ 48.6} \\ \hline 
   \textbf{door} &  \textbf{drawer} &  \textbf{open} & \textbf{closed} &1.5 \slash  1.0  &\textcolor{red}{ 1.5} \slash \textcolor{red}{ 0.6}  & 19.8 \slash  10.5  & 16.7 \slash  7.6  & 61.3 \slash  39.5  &\textcolor{blue}{ 67.3} \slash \textcolor{blue}{ 39.9} \\ \hline 
   \textbf{glass} &  \textbf{mug} &  \textbf{empty} & \textbf{CL} &1.1 \slash  0.7  &\textcolor{red}{ 0.8} \slash \textcolor{red}{ 0.6}  & 74.8 \slash  52.6  & 72.1 \slash  56.1  & 54.1 \slash  38.4  &\textcolor{blue}{ 85.3} \slash \textcolor{blue}{ 56.5} \\ \hline 
   \textbf{glass} &  \textbf{mug} &  \textbf{empty} & \textbf{CS} &3.3 \slash  2.4  &\textcolor{red}{ 2.9} \slash \textcolor{red}{ 2.2}  & 55.4 \slash  42.1  & 43.7 \slash  31.6  & 61.8 \slash  43.7  &\textcolor{blue}{ 78.4} \slash \textcolor{blue}{ 53.8} \\ \hline \hline 

\multicolumn{4}{|c||}{\textbf{Weighted Average mAP}}  &  8.5    & \textcolor{red}{6.4   }& 53.3  &   49.0    &   48.0 		 &   \color{blue} 72.5 \\ \hline
 \multicolumn{4}{|c||}{\textbf{Best perf (\%)  }}& \color{red}0.0   &  \color{red}0.0 &   \color{red}0.0 &   \color{red}0.0 &  14.3 & \color{blue}85.7 \\   \hline 
 \multicolumn{4}{|c||}{ \textbf{Worst perf. (\%)}} &  42.9&   \color{red}57.1 & \color{blue}0.0 & \color{blue}0.0 & \color{blue}0.0 & \color{blue}0.0  \\   \hline  \hline
  
     \multicolumn{4}{|c||}{\textbf{Weighted Average AP@50:5:95}} &  5.8   & \color{red}4.0 &  29.2 & 32.5 & 31.9 & \color{blue}45.1 \\   \hline 
  
   \multicolumn{4}{|c||}{\textbf{Best perf. (\%)}} &  \color{red}0.0   &  \color{red}0.0 &   14.3 &   \color{red}0.0 &  14.3 & \color{blue}71.4 \\   \hline 
 \multicolumn{4}{|c||}{ \textbf{Worst perf. (\%)}} &  42.9&   \color{red}57.1 & \color{blue}0.0 & \color{blue}0.0 & \color{blue}0.0 & \color{blue}0.0  \\   \hline 
\end{tabular}

\end{minipage}}
   \caption{ Experimental results for the TOOS Scenario.  Each row corresponds to a different objects/states combinations.  First  number in each cell corresponds to the mAP  metric and second number corresponds to the    AP@50:5:95 metric respectively. 
 Blue/red font indicates best/worst performance achieved for the given objects/states combination.}
    \label{table:ant}
\end{table*}

\section{Experimental Evaluation}

OD has been studied extensively and it is one of the most prominent cases where the standard deep learning approach, i.e. the training of special architectures of  Deep Neural Networks with appropriately annotated datasets, has been proven exceptionally advantageous in comparison to traditional techniques involving hand-crafted image features. We choose to follow the same approach for solving SD, since it allows to assess our intuition regarding the nature of OD and SD, both directly and indirectly. Specifically,  we can observe how the performance of SD variants changes as the complexity of the problem at hand varies, and check whether this behavior is congruent to the one that is expected  for OD variants when undergo a similar complexity shift. Equally importantly,  by testing SD under the same experimental conditions (i.e., identical training samples and network architectures) that were used for an OD problem,  we can compare directly the performances obtained for the two tasks.


\vspace*{0.2cm}\noindent\textbf{Training configuration:}
The network  we are using in our experimental evaluation is Yolov4~\cite{bochkovskiy2020yolov4}, one of the most popular CNNs for OD. We opted for using this network because it it can be easily fine-tuned, it has been utilized with much success in a wide variety of settings and exhibits SoA performance in OD. Consequently, we follow the typical procedure for the deployment of a network tailored for OD, i.e.,  we split the set of the annotated image samples into training, validation and testing parts. 

In order to accelerate the training phase and boost the performance, we follow the common practice of using  pre-trained weights at the start of the training phase of every  network. Specifically, the weights used are obtained after training with MS-COCO dataset~\cite{lin2014microsoft}. 

We also employ data augmentation by horizontally flipping all frames in a batch with a probability of 0.5.
The number of training epochs is set equal to the product of 2,000 to the number of the categories of the corresponding sub-task\footnote{{https://github.com/AlexeyAB/darknet\#custom-object-detection}}. We used the Adam optimizer with an initial learning rate of 0.001 and an early stopping policy of 50 epochs.

\vspace*{0.2cm}\noindent\textbf{Evaluation metrics:}
Given an image as an input, the solution of the SD problem produces the bounding box of each detected object and the set of state classes, in which that object belongs. For a detection to be considered correct, the set of states must be identical to the ground truth  and the overlap between the ground truth and the detected bounding boxes must be
 greater than a certain threshold, which, following common practices, we set to 50\%. Moreover,  predictions having  confidence lower than 50\% are rejected. Our analyses are based on three standard performance metrics used for assessing OD tasks: AP, AP@50:5:95 and mAP. AP is calculated for each object-state combination in each experiment instance, whereas AP@50:5:95 and mAP are calculated per experiment instance as in~\cite{padilla2020survey}. 

\noindent\textbf{Experimental scenarios:}
For the purposes of the evaluation, we devise 7 different scenarios, each one involving different assumptions regarding the object categories and states which are to be detected. 
The details of the scenarios are presented in Table~\ref{tab:exprs}.
In the rest of this section, we describe each scenario, we explain the rationale behind it along with the hypotheses that we want to test,  and present the experimental results we obtained and the conclusions that we draw.


\subsection{The OOOS, MOOS, OOMS and MOMS scenarios} 
Aiming to explore how the approach fares as the complexity of the problem scales, we devise 4 different SD experimental scenarios. The adjustable settings for the scenarios concern the number of different object categories involved and the number of possible states for each object category. Specifically, the tested SD scenarios are defined as follows:
\begin{itemize}
\item \textbf{OOOS (One Object One State pair)}: Involves the detection of an object's state, assuming the  object class is known, i.e. the corresponding network has been trained exclusively on objects belonging to this class, and the possible states for the object are two and mutually exclusive (M.E.).
\item  \textbf{MOOS (Many Objects One State pair):} involves the detection of an object's state  when its class is not known, i.e. the number of object classes   upon which the corresponding network has been trained is more than one, and the possible states for the objects are two and M.E.
\item \textbf{OOMS (One Object Many States):} involves the detection of an object's state(s) assuming the  object class is known and the possible states for the object are more than two and not necessarily M.E. 
\item \textbf{MOMS (Many Objects  Many States):} Detection of an object's state  when  its class is not known beforehand and the possible states for each class of object could be more than two and not necessarily M.E.
\end{itemize}

The obtained results are summarized in Tables~\ref{tab:table_aggr}-\ref{table:all2}. It can be verified that the best performance is obtained in the OOOS scenario, while the worst in the MOMS scenario. This is expected, as from a single object and state pair (OOOS) we move into the far more complex scenario of multiple objects and multiple state pairs (MOMS). However, we also observe that performances in MOOS are considerably better that performances in OOMS. Thus, when departing from the baseline OOOS scenario and increasing the complexity of the problem either in the direction of adding more objects (MOOS) or in the direction of adding more states (OOMS), we observe that the addition of more states makes the problem considerably harder.  

 \begin{table}[t]
    \centering
    \begin{tabular}{|l|c|c|}
    \hline
               & \textbf{State 1} & \textbf{State 2}   \\  \hline\hline

     \textbf{Object 1}  & \textbf{(A)} $O_1,S_1$ & \textbf{(B)} $O_1,S_2$  \\ \hline
     \textbf{Object 2}  & \textbf{(C)} $O_2,S_1$ & \textbf{(D)} $O_2,S_2$  \\ \hline
\end{tabular}

   \caption{Object/state combinations in the TOOS scenario. See text for details.}

    \label{tab:TOOS}
\end{table}

 \begin{table}[t]
    \centering
    
    \resizebox{0.86\textwidth}{!}{\begin{minipage}{\textwidth}
    \begin{tabular}{|l|l|l|c|}
    \hline
       \textbf{Setting}        & \textbf{Training} & \textbf{Testing} & \textbf{Task}  \\  \hline\hline

     Cross States 1 \textbf{(CS1)}   & A, D  & B, C & SD  \\ \hline
     Cross States 2 \textbf{(CS2)}   & B, C  & A, D & SD  \\ \hline
     States \textbf{(ST)}           & A, B, C, D & A, B, C, D & SD  \\ \hline \hline
      
     Cross Objects 1 \textbf{(CO1)}   & A, D & B, C & OD  \\ \hline 
     Cross Objects 2 \textbf{(CO2)}   & B, C & A, D & OD  \\ \hline
     Objects \textbf{(OB)}   & A, B, C, D & A, B, C, D & OD  \\ \hline
      
\end{tabular}
\end{minipage}}

   \caption{Dataset splits in each setting of the TOOS scenario. A, B, C and D are defined in Table~\ref{tab:TOOS}. See text for details.}

    \label{tab:TOOS2}
\end{table}

\begin{table*}[h]
    \centering
   
   \begin{tabular}{|l|c|c|c|c|c|c|c|c|} \hline   
  \diagbox[innerleftsep=.07cm,innerrightsep=.05pt]{\textbf{Trained}}{\textbf{Tested}}  
 
& \textbf{Book} & \textbf{Bottle} & \textbf{Box} & \textbf{Door} & \textbf{Drawer} & \textbf{Jar} & \textbf{Tub} \\ \hline \hline

\textbf{book} &  \textcolor{blue}{ 64.2}  \slash \textcolor{blue}{ 44.3} &  1.2  \slash  0.6 &  22.3  \slash  16.7 &  3.4  \slash  2.0 & \textcolor{red}{ 1.1}  \slash \textcolor{red}{ 0.5} &  2.3  \slash  1.4 &  9.9  \slash  5.4\\ \hline

\textbf{bottle} &   11.7  \slash  6.9 & \textcolor{blue}{ 71.5}  \slash \textcolor{blue}{ 48.0} &  5.6  \slash  3.8 & \textcolor{red}{\textcolor{red}{ 0.0}}  \slash \textcolor{red}{\textcolor{red}{ 0.0}} &  0.9  \slash  0.8 &  46.3  \slash  30.4 &  14.1  \slash  9.3\\ \hline

\textbf{box} &   35.1  \slash  22.8 & \textcolor{red}{ 4.1}  \slash  2.7 & \textcolor{blue}{ 73.0}  \slash \textcolor{blue}{ 53.4} &  5.1  \slash \textcolor{red}{ 1.7} &  28.9  \slash  13.6 &  11.0  \slash  6.7 &  59.5  \slash  42.7\\ \hline

\textbf{door} &   2.6  \slash  1.6 &  0.1  \slash \textcolor{red}{\textcolor{red}{ 0.0}} &  4.0  \slash  1.6 & \textcolor{blue}{ 42.1}  \slash \textcolor{blue}{ 23.9} &  5.1  \slash  1.5 & \textcolor{red}{\textcolor{red}{ 0.0}}  \slash \textcolor{red}{\textcolor{red}{ 0.0}} &  0.7  \slash  0.5\\ \hline

\textbf{drawer} &   17.4  \slash  7.4 &  0.1  \slash \textcolor{red}{\textcolor{red}{ 0.0}} &  9.8  \slash  4.6 &  8.2  \slash  5.0 & \textcolor{blue}{ 75.6}  \slash \textcolor{blue}{ 43.9} & \textcolor{red}{\textcolor{red}{ 0.0}}  \slash \textcolor{red}{\textcolor{red}{ 0.0}} & \textcolor{red}{\textcolor{red}{ 0.0}}  \slash \textcolor{red}{\textcolor{red}{ 0.0}}\\ \hline

\textbf{jar} &   1.7  \slash  1.0 &  14.0  \slash  7.1 &  1.9  \slash  1.1 & \textcolor{red}{\textcolor{red}{ 0.0}}  \slash \textcolor{red}{\textcolor{red}{ 0.0}} &  0.4  \slash  0.1 & \textcolor{blue}{ 50.0}  \slash \textcolor{blue}{ 25.8} &  21.2  \slash  11.0\\ \hline

\textbf{tub} &   4.8  \slash  2.4 &  2.1  \slash  1.3 &  15.9  \slash  10.6 & \textcolor{red}{ 0.2}  \slash \textcolor{red}{ 0.1} &  4.6  \slash  2.0 &  26.7  \slash  19.0 & \textcolor{blue}{ 83.0}  \slash \textcolor{blue}{ 59.1}\\ \hline

\end{tabular}

\caption{Experimental results  for the SDG scenario, for the mutually exclusive states pair P1. The first number in each cell corresponds to the mAP metric and the second to the AP@50:5:95  metric respectively.}
    \label{tab:exprSDGS1}
\end{table*}

\begin{table*}[thbp]
    \centering
      \footnotesize
   \begin{tabular}{|l|c|c|c|c|c|c|c|c|} \hline   
  \diagbox[innerleftsep=.07cm,innerrightsep=.05pt]{\textbf{Trained}}{\textbf{Tested}}  & \textbf{bottle} &\textbf{bowl} &\textbf{cup} &\textbf{glass} &\textbf{jar} &\textbf{mug} \\ \hline  
\textbf{bottle} &  \textcolor{blue}{ 38.0}  \slash \textcolor{blue}{ 26.3} & \textcolor{red}{ 1.6}  \slash \textcolor{red}{ 0.9} &  7.0  \slash  4.2 &  5.7  \slash  3.3 &  19.1  \slash  11.3 &  2.0  \slash  1.4\\ \hline

\textbf{bowl} &  \textcolor{red}{ 0.5}  \slash \textcolor{red}{ 0.4} & \textcolor{blue}{ 57.0}  \slash \textcolor{blue}{ 40.6} &  35.4  \slash  25.5 &  13.1  \slash  9.2 &  9.6  \slash  6.9 &  29.9  \slash  17.2\\ \hline

\textbf{cup} &  \textcolor{red}{ 1.1}  \slash \textcolor{red}{ 0.8} &  46.8  \slash  33.1 &  59.9  \slash  41.6 &  34.9  \slash  25.5 &  25.3  \slash  17.7 & \textcolor{blue}{ 63.3}  \slash \textcolor{blue}{ 47.5}\\ \hline

\textbf{glass} &  \textcolor{red}{ 1.3}  \slash \textcolor{red}{ 0.8} &  20.6  \slash  15.9 &  33.5  \slash  25.3 & \textcolor{blue}{ 60.1}  \slash \textcolor{blue}{ 41.5} &  19.7  \slash  14.9 &  14.4  \slash  7.7\\ \hline

\textbf{jar} &   10.7  \slash  5.1 & \textcolor{red}{ 2.4}  \slash \textcolor{red}{ 1.2} &  4.0  \slash  2.1 &  26.3  \slash  13.6 & \textcolor{blue}{ 35.8}  \slash \textcolor{blue}{ 25.0} &  5.6  \slash  2.6\\ \hline

\textbf{mug} &  \textcolor{red}{ 0.4}  \slash \textcolor{red}{ 0.2} &  46.3  \slash  32.8 &  52.1  \slash  35.7 &  31.3  \slash  20.2 &  15.4  \slash  10.0 & \textcolor{blue}{ 56.1}  \slash \textcolor{blue}{ 36.6}\\ \hline

\end{tabular}

\caption{Experimental results  for the SDG scenario, for the mutually exclusive states pair P2. The first number in each cell corresponds to the mAP metric and the second to the AP@50:5:95  metric respectively. Blue/red font indicates best/worst performance in a line.}
    \label{tab:exprSDGS2}
\end{table*}

\begin{table*}[h]
   
   \resizebox{0.80\textwidth}{!}{\begin{minipage}{\textwidth} 
   
    \centering
   
   \begin{tabular}{|l|c|c|c|c|c|c|c|c|c|c|} \hline   
  \diagbox[innerleftsep=.09cm,innerrightsep=.05pt]{\textbf{Trained}}{\textbf{Tested}}  & \textbf{basket} &\textbf{bottle} &\textbf{bowl} &\textbf{box} &\textbf{cup} &\textbf{glass} &\textbf{jar} &\textbf{mug} &\textbf{tub} \\ \hline   \hline  
\textbf{basket} &  \textcolor{blue}{ 55.6}  \slash \textcolor{blue}{ 29.1} &  17.4  \slash  10.0 &  22.4  \slash  15.7 &  29.9  \slash  20.5 &  17.4  \slash  12.6 & \textcolor{red}{ 10.2}  \slash \textcolor{red}{ 5.8} &  10.3  \slash  7.4 &  16.2  \slash  10.4 &  23.6  \slash  13.2\\ \hline

\textbf{bottle} &   5.5  \slash  3.4 & \textcolor{blue}{ 68.6}  \slash \textcolor{blue}{ 43.8} &  15.3  \slash  8.9 & \textcolor{red}{ 4.0}  \slash \textcolor{red}{ 2.6} &  32.9  \slash  22.3 &  31.0  \slash  20.1 &  41.4  \slash  27.6 &  11.9  \slash  6.4 &  17.9  \slash  10.1\\ \hline

\textbf{bowl} &   46.1  \slash  34.8 &  8.7  \slash  4.8 & \textcolor{blue}{ 74.9}  \slash \textcolor{blue}{ 52.4} & \textcolor{red}{ 5.7}  \slash \textcolor{red}{ 2.3} &  49.7  \slash  37.7 &  16.6  \slash  10.4 &  13.9  \slash  8.9 &  32.1  \slash  17.9 &  30.1  \slash  20.1\\ \hline

\textbf{box} &   42.2  \slash  21.3 &  17.5  \slash  9.1 &  42.8  \slash  18.9 & \textcolor{blue}{ 67.2}  \slash \textcolor{blue}{ 43.8} &  47.2  \slash  22.0 &  18.9  \slash  9.5 &  20.5  \slash  10.4 & \textcolor{red}{ 17.2}  \slash \textcolor{red}{ 9.0} &  39.4  \slash  18.5\\ \hline

\textbf{cup} &   37.1  \slash  23.5 & \textcolor{red}{ 7.4}  \slash  5.1 &  47.7  \slash  31.2 &  8.2  \slash \textcolor{red}{ 4.2} &  62.2  \slash \textcolor{blue}{ 47.8} &  43.0  \slash  29.1 &  21.4  \slash  15.1 & \textcolor{blue}{ 65.0}  \slash  41.5 &  19.8  \slash  13.1\\ \hline

\textbf{glass} &   17.4  \slash  10.4 &  10.4  \slash  4.6 &  25.7  \slash  15.2 & \textcolor{red}{ 3.1}  \slash \textcolor{red}{ 1.5} &  41.4  \slash  25.0 & \textcolor{blue}{ 57.9}  \slash \textcolor{blue}{ 41.0} &  17.9  \slash  10.1 &  24.5  \slash  12.7 &  11.9  \slash  7.1\\ \hline

\textbf{jar} &  \textcolor{red}{ 9.2}  \slash \textcolor{red}{ 5.5} &  24.0  \slash  16.4 &  21.1  \slash  15.3 &  15.0  \slash  9.4 &  22.5  \slash  14.7 &  22.0  \slash  14.8 & \textcolor{blue}{ 51.8}  \slash \textcolor{blue}{ 33.4} &  14.1  \slash  9.7 &  20.5  \slash  11.7\\ \hline

\textbf{mug} &   16.2  \slash  11.3 &  11.6  \slash  7.3 &  55.1  \slash  37.9 & \textcolor{red}{ 0.7}  \slash \textcolor{red}{ 0.6} &  61.0  \slash  42.9 &  28.9  \slash  19.3 &  12.3  \slash  8.2 & \textcolor{blue}{ 64.6}  \slash \textcolor{blue}{ 45.8} &  15.6  \slash  10.1\\ \hline

\textbf{tub} &   31.8  \slash  18.5 &  27.2  \slash  15.8 & \textcolor{blue}{ 64.9}  \slash  38.2 & \textcolor{red}{ 19.5}  \slash \textcolor{red}{ 10.1} &  44.6  \slash  30.8 &  21.8  \slash  12.6 &  31.1  \slash  18.4 &  27.2  \slash  14.5 &  64.2  \slash \textcolor{blue}{ 39.7}\\ \hline

\end{tabular}
\end{minipage}}
\caption{Experimental results  for the SDG scenario, for the mutually exclusive states pair P3. The first number in each cell corresponds to the mAP metric and the second to the AP@50:5:95  metric respectively. Blue/red font indicates best/worst performance in a line.}
    \label{tab:exprSDGS3}
\end{table*}

 \begin{table*}[h]

   \centering
   
   	\footnotesize
 
  \begin{tabular}{|*{4}{c|}|*{4}{c|}} \hline   
   \diagbox[innerleftsep=.07cm,innerrightsep=.05pt]{\textbf{Trained}}{\textbf{Tested}} & \textbf{charger} &\textbf{phone} & \textbf{socket} &

   \diagbox[innerleftsep=.07cm,innerrightsep=.05pt]{\textbf{Trained}}{\textbf{Tested}} & \textbf{newspaper}  & \textbf{shirt}  & \textbf{towel} \\ \hline

 \textbf{charger} &  \textcolor{blue}{ 68.6}  \slash \textcolor{blue}{ 42.2} &  12.6  \slash  7.8 & \textcolor{red}{ 3.1}  \slash \textcolor{red}{ 0.7}  &
\textbf{newspaper} &  \textcolor{blue}{ 35.7}  \slash \textcolor{blue}{ 23.5} & \textcolor{red}{ 6.9}  \slash \textcolor{red}{ 4.3} &  12.2  \slash  8.0\\ \hline

\textbf{phone} &   22.4  \slash  9.5 & \textcolor{blue}{ 61.7}  \slash \textcolor{blue}{ 40.6} & \textcolor{red}{\textcolor{red}{ 0.0}}  \slash \textcolor{red}{\textcolor{red}{ 0.0}} &
 
\textbf{shirt} &  \textcolor{red}{ 11.0}  \slash \textcolor{red}{ 5.1} & \textcolor{blue}{ 41.8}  \slash \textcolor{blue}{ 22.5} &  25.5  \slash  18.5\\ \hline

\textbf{socket} &   4.0  \slash  0.9 & \textcolor{red}{\textcolor{red}{ 0.0}}  \slash \textcolor{red}{\textcolor{red}{ 0.0}} & \textcolor{blue}{ 76.2}  \slash \textcolor{blue}{ 39.3}  &

\textbf{towel} &  \textcolor{red}{ 19.7}  \slash \textcolor{red}{ 12.7} &  34.0  \slash  23.5 & \textcolor{blue}{ 53.9}  \slash \textcolor{blue}{ 35.5}\\ \hline

 \end{tabular}

   \caption{Experimental results  for the SDG scenario, for the mutually exclusive states pair P4 (left table) and P5 (right table) respectively. The first number in each cell corresponds to the mAP metric and the second to the AP@50:5:95  metric respectively. Blue/red font indicates best/worst performance in a line}
  \label{tab:exprSDGS4}
 \end{table*}
\subsection{Object Detection Scenario (ODS) vs MOMS}
The  Object Detection Scenario (ODS) deals with the detection of  object categories, using the same settings used for the MOMS scenario. Essentially, it is about the same data, processed with the same network architectures, with the exception that in ODS we seek for the object categories, while in MOMS we seek for the object states. 

The performances of the detectors for the two scenarios are presented in Table~\ref{tab:exprMOOS}. We observe that the performance in OD is better than that achieved in SD. The better ODS performance is obtained despite the fact that ODS has to deal with double the number of categories (18) compared to MOMS (9).
Furthermore, the existence of visually similar pairs of objects (bottle-jar, cup-glass, towel-shirt) increase the difficulty of the OD task. Overall, the comparison between the performances of the two scenarios  supports strongly the notion that SD is a harder problem to solve than OD.


 
    
      
 
  
  
  
  

\subsection{The Two Objects One State pair (TOOS) scenario} 
In the context of the Two Objects One State Pair (TOOS) scenario, we train and test for images in which  there are two different types of objects appearing in either one of two mutually exclusively states. Specifically, let $O_1$ and $O_2$ be the two objects and $S_1$ and $S_2$ the two states. The situation is illustrated in Table~\ref{tab:TOOS2}. We define six different cases, depending on which subset of data is used for training, which for testing and what task is solved. 
All these splits have been employed in 12 different object/states combinations.
The rationale behind these experiments  is to compare directly the learning capacity of the detectors for object classes and object categories, when trained and tested on exactly the same base of data.

Table~\ref{table:ant}
 summarizes the obtained results. The performance in CS1 and CS2 (aiming for states) is significantly inferior to  the one observed in cases CO1 and CO2 (aiming for objects). The poor performance in this case can be explained by the fact that the detectors learn principally the visual features of the object category and not the ones of the state class. In other words, the category of an object is visually much more salient than its state. Additionally, the performance in ST is lower than the performance on OB. This evidence supports the hypothesis that SD is harder than OD. 






\subsection{State Detection Generalization (SDG) scenario}
In the State Detection Generalization Scenario (SDG), we examine the generalization capacity of the state detectors. Specifically, in the context of this scenario, state detectors which have been trained for instances of one particular object category and a  pair of two mutually exclusive states are tested on instances of other categories of objects that appear on either one of the same two mutually exclusively states. As an example, we check whether an SD method trained on bottles to detect whether they are filled or empty, is used to check whether a glass is filled or empty. In this case, we expect that the detectors will be able to generalize better for objects that are visually similar to the object class for the states of which they have been trained.

The results for this scenario for the pair P1 of states Open/Close, is shown in Tables~\ref{tab:exprSDGS1}-\ref{tab:exprSDGS4}.
The obtained results corroborate the intuition that the performance of a state detector increases along  with the visual similarity between the objects classes of the trained and the tested samples. In more detail, the performances are high for specific cases of great visual similarity and mediocre or poor for the rest of the cases. The fact that the state detectors can only generalize for certain cases makes probable that the state detectors will  exhibit poor performance  when encountering new types of objects. The above conclusions are consistent with the ones obtained from all the rest experiments performed on the rest of mutually exclusive pairs of states (P2-P5), the results of which are reported in the supplementary material.

\section{Summary and Future Work}







In this paper, we have introduced a new dataset of object states and conducted an extensive series of experiments over it in order to compare the SD and OD tasks. 
Overall,  the experimental results indicate the significant differences of the SD and OD tasks. 
SD is more challenging than OD. Moreover, the results obtained for more than a single pair of  state classes can be considered mediocre to poor. In this case, better performance could be obtained if a number of OOOS or MOOS detectors are employed simultaneously but the scalability issues of this approach limit its practical use. 

Regarding future work, there are a number of steps worth exploring. First, a more fine-grained categorization of the states would be more convenient especially for SDs  used in the context of AR. Moreover, an experimental evaluation using a two-stage SoA object detector, such as Faster-RCNN~\cite{ren2015faster}, could provide additional insights with respect to the similarities and differences between SD and OD. The utilization of semantic embeddings, which have been used with great success in  challenging problems such as Zero-Shot recognition, is another avenue that seems promising. Another interesting avenue of research would be to examine how the grouping of objects according to common sense criteria affects the performance of the state detectors. Obviously, the main goal is to come up with a novel method for SD which will cope with this important and  difficult problem in realistic conditions.


\bibliographystyle{apalike}
{\small
\bibliography{main}}

\begin{thebibliography}{}

\bibitem[Aboubakr et~al., 2019]{Aboubakr2019}
Aboubakr, N., Crowley, J.~L., and Ronfard, R. (2019).
\newblock {Recognizing Manipulation Actions from State-Transformations}.

\bibitem[Alayrac et~al., 2017]{Alayrac17}
Alayrac, J.-B., Sivic, J., Laptev, I., and Lacoste-Julien, S. (2017).
\newblock {Joint Discovery of Object States and Manipulation Actions}.
\newblock In {\em International Conference on Computer Vision (ICCV)}.

\bibitem[Bertasius and Torresani, 2020]{Bertasius2020}
Bertasius, G. and Torresani, L. (2020).
\newblock {COBE: Contextualized Object Embeddings from Narrated Instructional
  Video}.
\newblock pages 1--14.

\bibitem[Bochkovskiy et~al., 2020]{bochkovskiy2020yolov4}
Bochkovskiy, A., Wang, C.-Y., and Liao, H.-Y.~M. (2020).
\newblock Yolov4: Optimal speed and accuracy of object detection.
\newblock {\em arXiv preprint arXiv:2004.10934}.

\bibitem[Fathi and Rehg, 2013]{Fathi2013}
Fathi, A. and Rehg, J.~M. (2013).
\newblock {Modeling actions through state changes}.
\newblock {\em Proceedings of the IEEE Computer Society Conference on Computer
  Vision and Pattern Recognition}, pages 2579--2586.

\bibitem[Fire and Zhu, 2015]{Fire2015}
Fire, A. and Zhu, S.~C. (2015).
\newblock {Learning perceptual causality from video}.
\newblock {\em ACM Transactions on Intelligent Systems and Technology}, 7(2).

\bibitem[Fire and Zhu, 2017]{fire2017inferring}
Fire, A. and Zhu, S.-C. (2017).
\newblock Inferring hidden statuses and actions in video by causal reasoning.
\newblock In {\em Proceedings of the IEEE Conference on Computer Vision and
  Pattern Recognition Workshops}, pages 48--56.

\bibitem[Girshick, 2015]{girshick2015fast}
Girshick, R. (2015).
\newblock Fast r-cnn.
\newblock In {\em Proceedings of the IEEE international conference on computer
  vision}, pages 1440--1448.

\bibitem[He et~al., 2017]{he2017mask}
He, K., Gkioxari, G., Doll{\'a}r, P., and Girshick, R. (2017).
\newblock Mask r-cnn.
\newblock In {\em Proceedings of the IEEE international conference on computer
  vision}, pages 2961--2969.

\bibitem[He et~al., 2016]{he2016deep}
He, K., Zhang, X., Ren, S., and Sun, J. (2016).
\newblock Deep residual learning for image recognition.
\newblock In {\em Proceedings of the IEEE conference on computer vision and
  pattern recognition}, pages 770--778.

\bibitem[Huang et~al., 2017]{huang2017densely}
Huang, G., Liu, Z., Van Der~Maaten, L., and Weinberger, K.~Q. (2017).
\newblock Densely connected convolutional networks.
\newblock In {\em Proceedings of the IEEE conference on computer vision and
  pattern recognition}, pages 4700--4708.

\bibitem[Isola et~al., 2015]{Isola2015}
Isola, P., Lim, J.~J., and Adelson, E.~H. (2015).
\newblock {Discovering states and transformations in image collections}.
\newblock {\em Proceedings of the IEEE Computer Society Conference on Computer
  Vision and Pattern Recognition}, 07-12-June:1383--1391.

\bibitem[Lin et~al., 2017a]{lin2017feature}
Lin, T.-Y., Doll{\'a}r, P., Girshick, R., He, K., Hariharan, B., and Belongie,
  S. (2017a).
\newblock Feature pyramid networks for object detection.
\newblock In {\em Proceedings of the IEEE conference on computer vision and
  pattern recognition}, pages 2117--2125.

\bibitem[Lin et~al., 2017b]{lin2017focal}
Lin, T.-Y., Goyal, P., Girshick, R., He, K., and Doll{\'a}r, P. (2017b).
\newblock Focal loss for dense object detection.
\newblock In {\em Proceedings of the IEEE international conference on computer
  vision}, pages 2980--2988.

\bibitem[Lin et~al., 2014]{lin2014microsoft}
Lin, T.-Y., Maire, M., Belongie, S., Hays, J., Perona, P., Ramanan, D.,
  Doll{\'a}r, P., and Zitnick, C.~L. (2014).
\newblock Microsoft coco: Common objects in context.
\newblock In {\em European conference on computer vision}, pages 740--755.
  Springer.

\bibitem[Liu et~al., 2016]{liu2016ssd}
Liu, W., Anguelov, D., Erhan, D., Szegedy, C., Reed, S., Fu, C.-Y., and Berg,
  A.~C. (2016).
\newblock Ssd: Single shot multibox detector.
\newblock In {\em European conference on computer vision}, pages 21--37.
  Springer.

\bibitem[Liu et~al., 2017]{Liu2017}
Liu, Y., Wei, P., and Zhu, S.~C. (2017).
\newblock {Jointly Recognizing Object Fluents and Tasks in Egocentric Videos}.
\newblock {\em Proceedings of the IEEE International Conference on Computer
  Vision}, 2017-Octob:2943--2951.

\bibitem[Mahdisoltani et~al., 2018]{mahdisoltani2018effectiveness}
Mahdisoltani, F., Berger, G., Gharbieh, W., Fleet, D., and Memisevic, R.
  (2018).
\newblock On the effectiveness of task granularity for transfer learning.
\newblock {\em arXiv preprint arXiv:1804.09235}.

\bibitem[Padilla et~al., 2020]{padilla2020survey}
Padilla, R., Netto, S.~L., and da~Silva, E.~A. (2020).
\newblock A survey on performance metrics for object-detection algorithms.
\newblock In {\em 2020 International Conference on Systems, Signals and Image
  Processing (IWSSIP)}, pages 237--242. IEEE.

\bibitem[Redmon and Farhadi, 2017]{redmon2017yolo9000}
Redmon, J. and Farhadi, A. (2017).
\newblock Yolo9000: better, faster, stronger.
\newblock In {\em Proceedings of the IEEE conference on computer vision and
  pattern recognition}, pages 7263--7271.

\bibitem[Ren et~al., 2015]{ren2015faster}
Ren, S., He, K., Girshick, R., and Sun, J. (2015).
\newblock Faster r-cnn: Towards real-time object detection with region proposal
  networks.
\newblock {\em Advances in neural information processing systems}, 28:91--99.

\bibitem[Wang et~al., 2016]{wang2016actions}
Wang, X., Farhadi, A., and Gupta, A. (2016).
\newblock Actions\~{} transformations.
\newblock In {\em Proceedings of the IEEE conference on Computer Vision and
  Pattern Recognition}, pages 2658--2667.

\end{thebibliography}

\end{document}